\documentclass[a4paper]{article}

\newcommand{\timePoint}[1]{{\sffamily \footnotesize({\small \textbf{t}$_{#1}$}\footnotesize)}}

\usepackage{graphicx}
\usepackage{onecolceurws}

\usepackage[T1]{fontenc}

\usepackage{hyperref}
\hypersetup{
    bookmarks=false,         
    unicode=false,          
    pdftoolbar=true,        
    pdfmenubar=true,        
    pdffitwindow=false,     
    pdfstartview={FitH},    
    pdftitle={Towards a Human-Centred Cognitive Model of Visuospatial Complexity in Everyday Driving},    
    pdfauthor={Vasiliki Kondyli, Mehul Bhatt, Jakob Suchan},     
    pdfsubject={},   
    pdfcreator={Mehul Bhatt},   
    pdfproducer={}, 
    pdfkeywords={}, 
    pdfnewwindow=true,      
    colorlinks=true,       
    linkcolor=blue!80!black,          
    citecolor=blue!80!black,        
    urlcolor=blue!80!black,           
    filecolor=black      
}

\usepackage{pbox}
\usepackage{cleveref}

\usepackage{subcaption}

\usepackage{booktabs}
\usepackage[english]{babel}
\usepackage{amssymb}
\usepackage{url}
\usepackage{xcolor}
\usepackage[export]{adjustbox}
\usepackage{colortbl}
\makeatletter
\def\hlinewd#1{%
  \noalign{\ifnum0=`}\fi\hrule \@height #1 \futurelet
   \reserved@a\@xhline}
\makeatother

\title{{\sffamily\Large Towards a Human-Centred Cognitive Model of\\Visuospatial Complexity in Everyday Driving}}

\author{
{\normalsize Vasiliki Kondyli}
\and
{\normalsize Mehul Bhatt}
\and
{\normalsize Jakob Suchan}
}

\institution{\"{O}rebro University, Sweden\\University of Bremen, Germany\\CoDesign Lab $~${\footnotesize>}$~$  Cognitive Vision\\{\small\href{http://www.codesign-lab.org/cognitive-vision}{www.codesign-lab.org/cognitive-vision}}}

\begin{document}
\maketitle


\begin{abstract}{\small\sffamily

We develop a human-centred cognitive model of visuospatial complexity in everyday, naturalistic driving conditions. With a focus on visual perception, the model incorporates quantitative, structural, and dynamic attributes identifiable in the chosen context; the human-centred basis of the model lies in its behavioural evaluation with human subjects with respect to psychophysical measures pertaining to embodied visuoauditory attention. We report preliminary steps to apply the developed cognitive model of visuospatial complexity for human-factors guided dataset creation and benchmarking, and for its use as a semantic template for the (explainable) computational analysis of visuospatial complexity.

}
\end{abstract}

\smallskip

\section{{\sffamily Introduction}}
Autonomous driving research has received enormous academic \& industrial interest in recent years. As the self-driving vehicle industry develops, it will be necessary ---in a manner similar to sectors such as medical computing, computer aided engineering and design--- to have an articulation and community consensus on aspects such as representation, interoperability, data archival \& retrieval mechanisms, and human-centred performance benchmarks. We posit that this will be necessary towards fulfilling essential legal and ethical responsibilities, such as those pertaining to representation and realisation of rules and norms, verifiable performance  of normative behaviour, and technology capabilities vis-a-vis human-centred expectations.

\smallskip

\textbf{\sffamily Human-Centred Benchmarking and Standardisation}.\quad Within autonomous driving, the need for standardisation and ethical regulation has most recently garnered interest internationally, e.g.,  with the Federal Ministry of Transport and Digital Infrastructure in Germany (BMVI) taking a lead in eliciting 20 key propositions (with possible legal implications) for the fulfilment of ethical commitments for automated and connected driving systems \cite{BMVI2018}. In spite of major investments in self-driving vehicle research, issues related to human-centred design, human-machine interaction, and standardisation have been barely addressed, with  the current focus in driving research primarily being on two basic considerations: \emph{how fast to drive, and which way and how much to steer}. This is necessary, but inadequate if autonomous vehicles are to become commonplace and function with humans:  not everything in autonomous vehicles is about realtime control/decision-making; several human-machine interaction requirements (e.g., for diagnostic communication, universal design) also exist. The  $20$ key propositions elicited by the German federal ministry BMVI highlight a range of factors pertaining to safety, utilitarian considerations, human rights, statutory liability, technological transparency, data management and privacy etc. Ethically driven standardisation and regulation will require addressing challenges in foundational human-centred AI technology, e.g., pertaining to semantic visual interpretation, natural / multimodal human-machine interaction, high-level data analytics (e.g., for post hoc diagnostics, dispute settlement) \cite{SuchanECAI20,SuchanIJCAI19}. This will necessitate --amongst other things-- human-centred qualitative benchmarks relevant to different facets of machine (visual) intelligence, and incorporation of multifaceted hybrid AI solutions to fulfil such requirements. We claim that what appears as a spectrum of complex challenges (in autonomous driving) are actually rooted to one fundamental methodological consideration that needs to be prioritised, namely: the design and implementation of human-centred technology based on a \emph{confluence} of techniques and perspectives from AI+ML, Cognitive Science \& Psychology, Human-Machine Interaction, and Design Science. Like in many applications of AI, such an integrative approach has so far not been mainstream also within autonomous driving research.

 \begin{figure}[t]
 \centering
 \begin{subfigure}[c]{0.21\textwidth}
 \includegraphics[width=\textwidth]{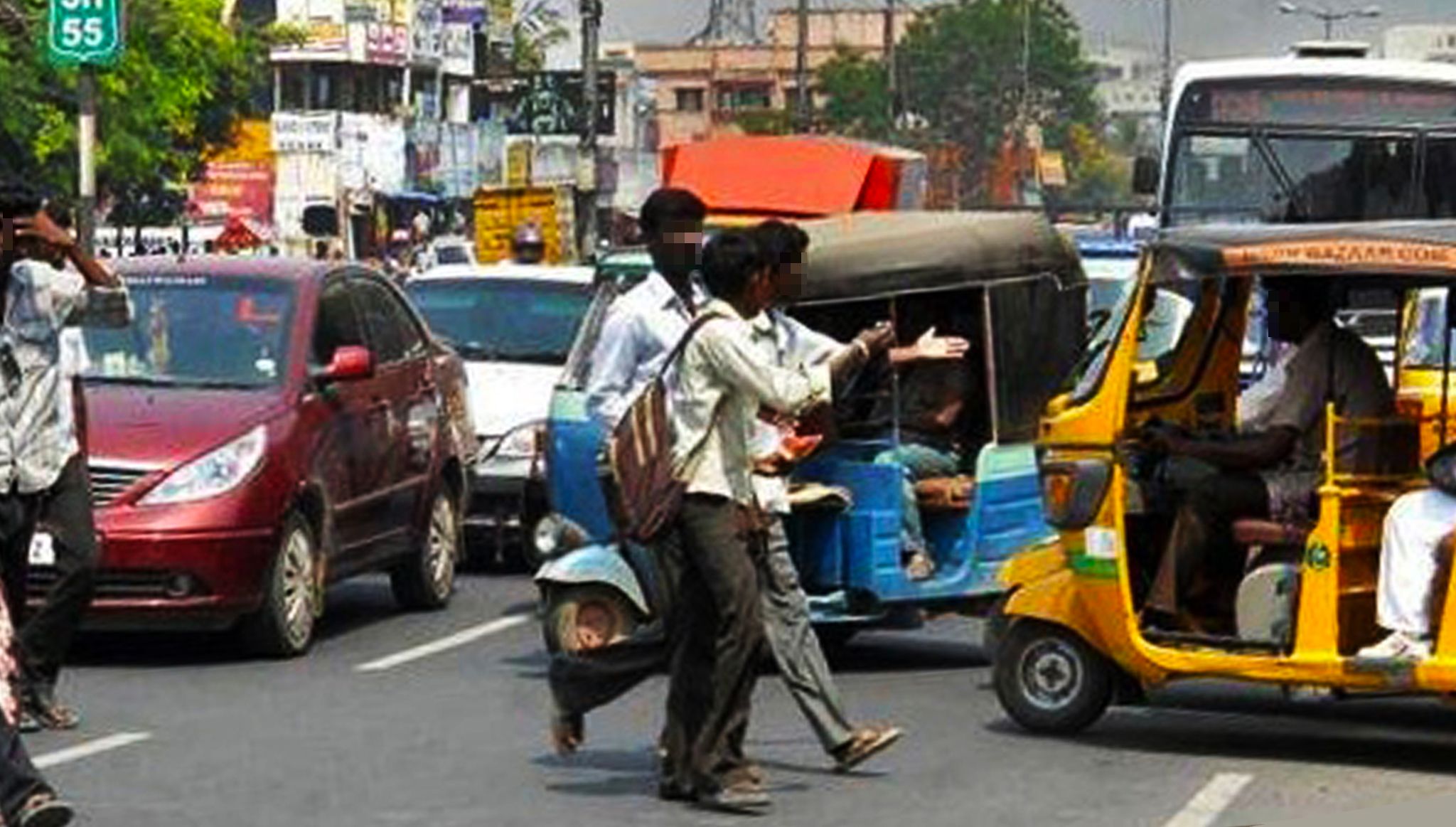}
 \subcaption{}
 \label{india_gesture}
 \end{subfigure}
 \begin{subfigure}[c]{0.21\textwidth}
 \includegraphics[width=\textwidth]{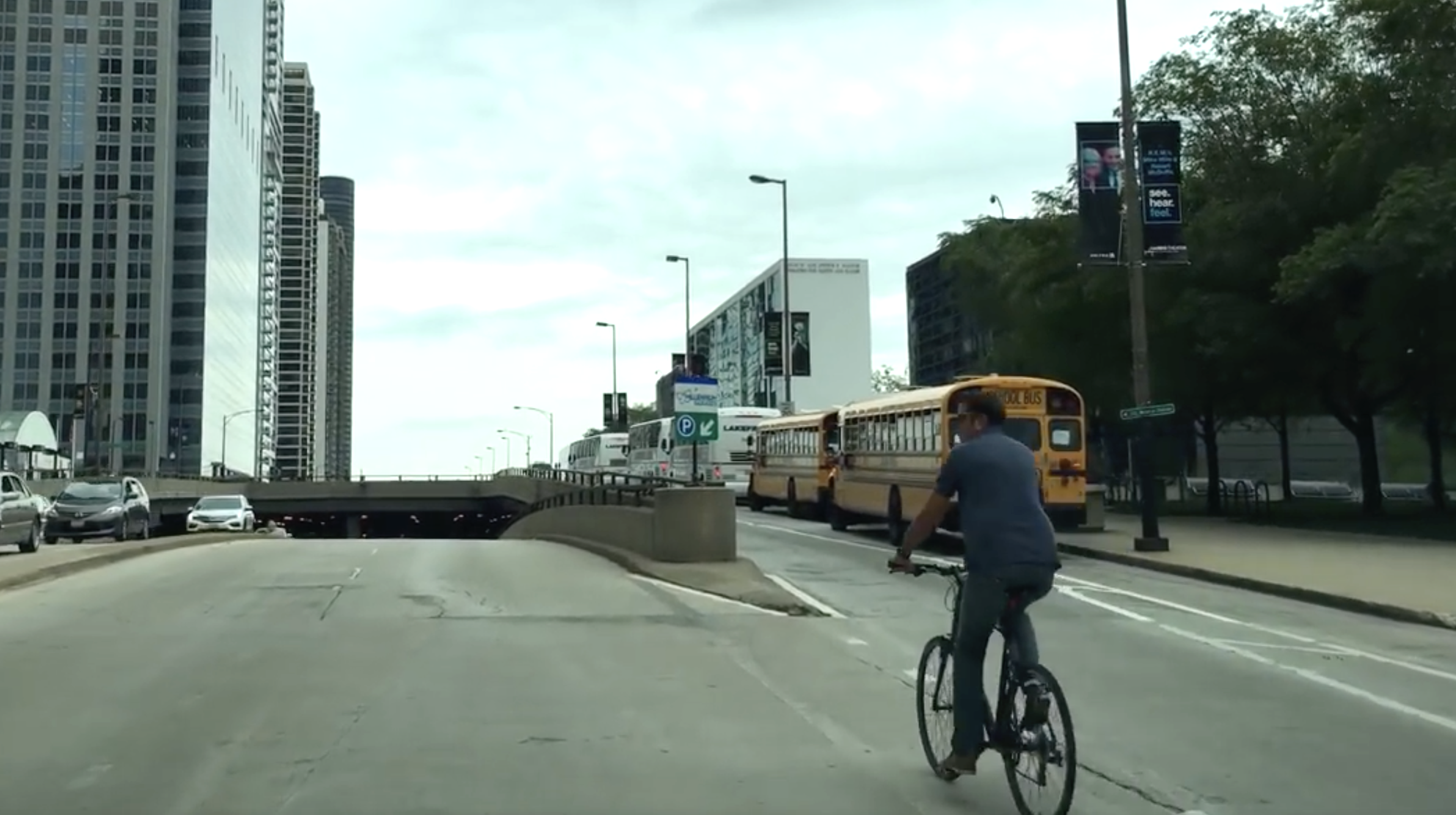}
 \subcaption{}
 \label{bike}
 \end{subfigure}
 \begin{subfigure}[c]{0.21\textwidth}
  \includegraphics[width=\textwidth]{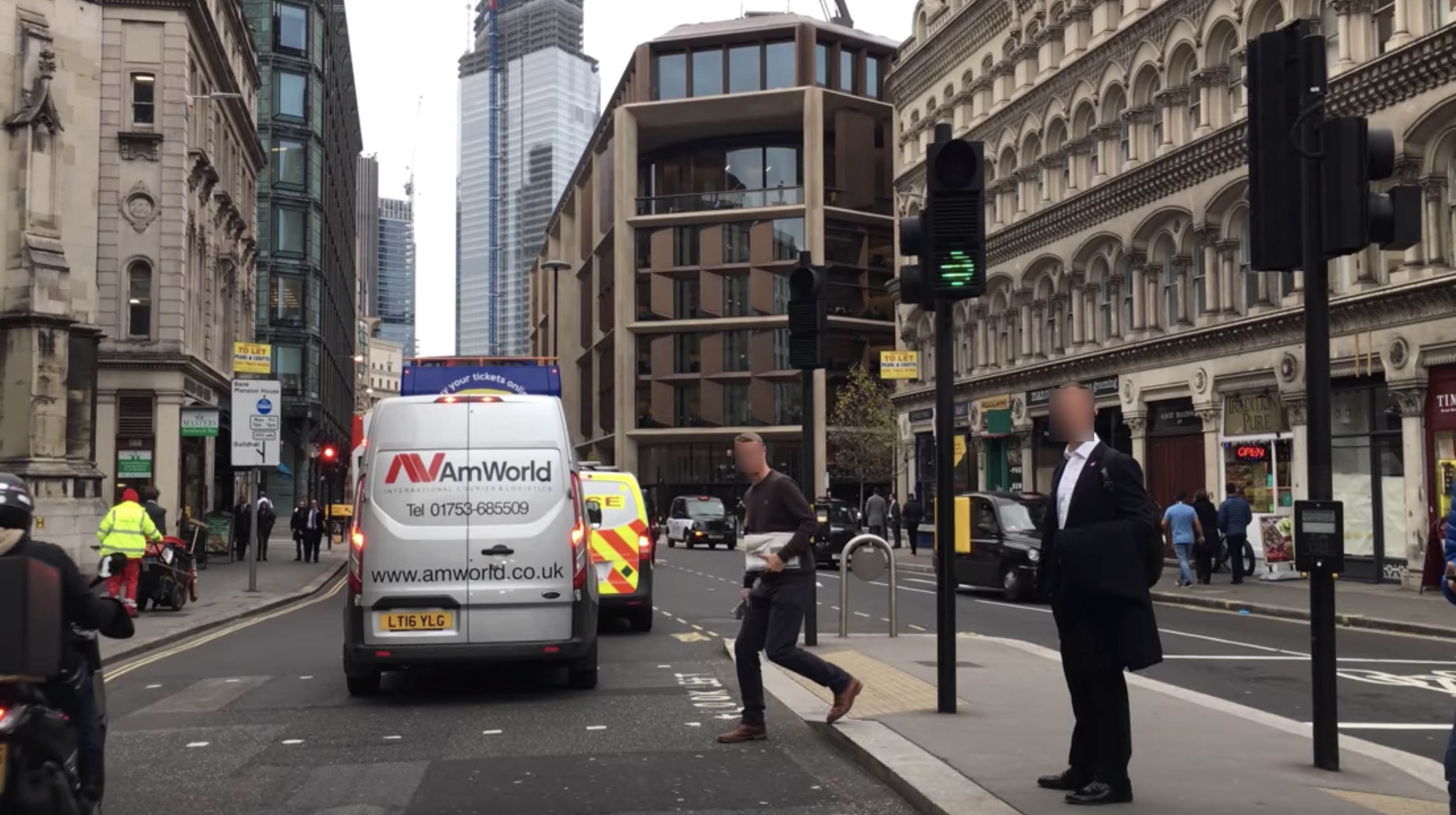}
 \subcaption{}
 \label{london_steponstreet}
 \end{subfigure}
 \begin{subfigure}[c]{0.21\textwidth}
 \includegraphics[width=\textwidth]{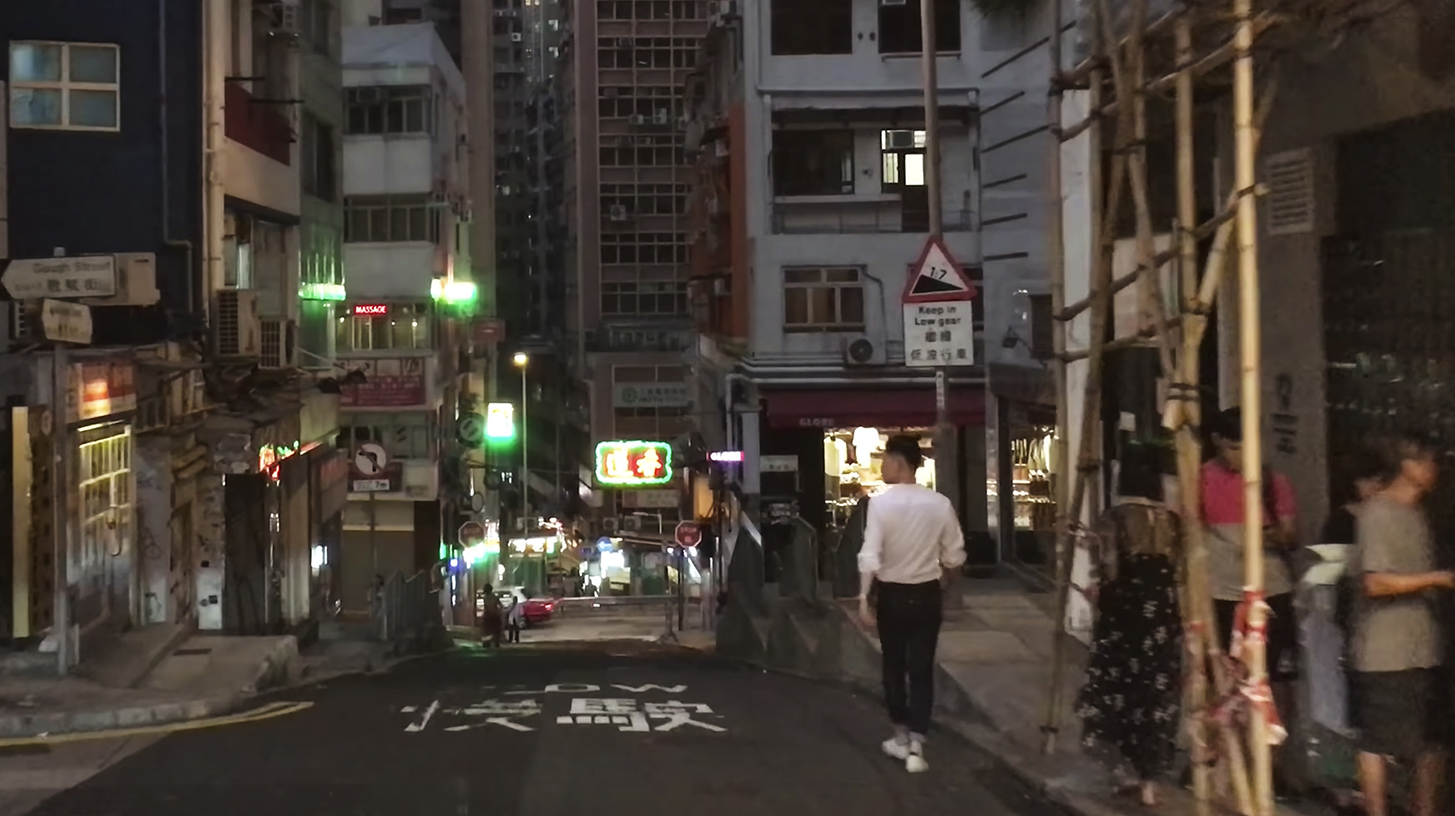}
 \subcaption{}
 \label{hongkong}
 \end{subfigure}
 
\caption{{\sffamily\small Multimodal communication in complex streetscape scenes:  \textbf{(a)} Pedestrian gesturing and head rotation (Mumbai); \textbf{(b)} Cyclist's head rotation (Chicago); \textbf{(c)} Pedestrian stepping to the street and monitoring attention of the driver (London); \textbf{(d)} Pedestrian moving to the side of the street (Hong Kong).}}
 \label{}
 \end{figure}

\smallskip

\textbf{\sffamily Multimodal Interaction: The Case of Visuospatial Complexity}.\quad The central focus of the research presented in this paper is to develop a systematic methodology for the development of human-centred benchmarks for visual sensemaking and multimodal interaction for the domain of autonomous vehicles. Our work is driven by a bottom-up interdisciplinary approach --combining techniques in AI, Psychology, HCI, and design-- for the study of embodied multimodal interaction in diverse ecologically valid conditions, with a particular emphasis on low-speed driving and complex urban environments (possibly also with unstructured traffic and people dynamics). We emphasise driver behaviour, as well as behaviour of other users such as pedestrians, bicyclists, bikers whilst focussing on natural interactions (e.g., gestures, joint attention) amongst involved stakeholders.

\smallskip

\textbf{\sffamily Key Contribution}.\quad We present a \emph{cognitive model of visuospatial complexity} in everyday driving situations that may be used as a basis to \emph{design, evaluate, standardise, and test \& validate}  visuospatial sensemaking capabilities of computational models of visual intelligence (for autonomous vehicles). We posit that our methodology can be used as a basis of developing human-centred benchmark datasets for visual sensemaking in autonomous driving encapsulating key cognitive principles founded on empirically established (context-specific) embodied multimodal interaction patterns under naturalistic driving conditions in everyday life \cite{angrosino2007naturalistic,social-interaction-ecologica-2016}. The proposed model of visuospatial complexity is based on \emph{quantitative, structural}, and \emph{dynamic } roadside attributes identifiable from the stimuli under consideration. As an example application of our model, we report work-in-progress concerning the development of one instance of a dataset where the central emphasis is on the evaluation of visuospatial complexity of driving stimuli. Both our proposed methodology and its human evaluation are driven by behavioural research in visual and spatial cognition methods as pursued within cognitive and environmental psychology. Such bottom-up interdisciplinary studies combining AI, psychology, HCI, and design are needed to better appreciate the complexity and spectrum of varied human-centred challenges in autonomous driving.

\section{\sffamily{{Naturalistic Human Factors in Everyday Driving}}}

We approach cognitively motivated human-factors in autonomous driving from the viewpoint of visual attention (e.g. visual search, change detection), spatial cognition (e.g. cognitive mapping), and multimodal communication (e.g. joint attention, gesturing). We conduct systematic behavioural studies with a representative set of scenarios that encompasses challenging, hazardous, or cognitively demanding events in the urban set-up. The scenarios involve a range of visuospatial complexity levels in streetscape that reflects different areas of the world. 
\smallskip

\textbf{\sffamily Multimodal Interactions (in the Streetscape)} \quad Multimodal interactions are highly varied and they can convey very different meanings depending on the task, the environmental setting, or other social, and dynamic factors. Even if they seem everyday monotonic events, they are complex problems for today's autonomous systems. Embodied interactions (e.g., on the street) use a combination of communication modalities such as gestures and speech to pass a message or solve a conflict (Fig. \ref{india_gesture}), or head rotations and eye contact to establish joint attention (Fig. \ref{bike})  (Table \ref{multimodal_interactions}). A practical action when observed by an agent can be considered both as a physical action as well as a message (e.g., a pedestrian stepping on the street indicating willingness to get priority; Fig. \ref{london_steponstreet}). Consequently, street users receive and comprehend different implicit cues that may alter their behaviour  based on their observations and their expectations of being observed (e.g. a pedestrian is moving to the side of the street based on the auditory cue of a car approaching and the estimation of street's width; Fig. \ref{hongkong}). 
 \smallskip

  {
 \begin{table}[t]
\renewcommand{\arraystretch}{1.2}
 \begin{center}
 \scriptsize\sffamily
\begin{tabular}{>{\columncolor[gray]{0.98}}l p{13.5cm}}
 \hlinewd{1pt}
\rowcolor[gray]{.92} \textbf{MODALITIES} & \textbf{Examples} \\\hline

SPEECH & Ask$~$-$~$Warn$~$-$~$Shout$~$-$~$Scold$~$-$~$Give directions \\
HEAD MOVEMENTS  & Turn towards the street$~$-$~$Tilt to a direction$~$-$~$Nod for disapproval$~$-$~$Slide for notice$~$-$~$Protrusion for warning \\
FACIAL EXPRESSIONS & Smiles$~$-$~$Frowns$~$-$~$Wrinkle$~$-$~$Eye Rolling$~$-$~$Cut Eye$~$-$~$Eyebrows Raising$~$-$~$Lips Movement$~$-$~$Mouth Movement\\
GESTURES & Emblematic (hitchhiking, stop)$~$-$~$Iconic (direction of movement)$~$-$~$Deictic (pointing)$~$-$~$Beat (irritation, gratitude)\\
BODY POSTURES  & Crossing arms$~$-$~$Idle$~$-$~$Stand with the back to the street$~$-$~$Lean towards$~$-$~$Stand besides a car/bike  \\
GAZE   & Eye contact$~$-$~$Seek attention$~$-$~$Follow other's gaze$~$-$~$Follow a moving object$~$-$~$Aversion$~$-$~$Point towards a direction \\
AUDITORY CUES  & Honking$~$-$~$Car engine$~$-$~$Traffic light sound$~$-$~$Brakes$~$-$~$Siren$~$-$~$Voice  \\
\hline
\rowcolor[gray]{.92} \textbf{PRACTICAL ACTIONS} & \textbf{(Select Sample)} \\\hline
 \multicolumn{2} {p{16.7cm}} {Cyclist standing beside the bike while putting on the helmet indicates he will start cycling soon$~$-$~$Pedestrian on a wheelchair beside the street  with    hands on the wheels indicates his intention to cross$~$-$~$Pedestrian pushing the button for the traffic light, not monitoring the traffic and relying on auditory cues$~$-$~$Driver switching the light on and off to indicate that the cyclist can cross, combined with a node to the direction of movement$~$-$~$ Driver  slowing down/accelerating indicates intentions to give / take priority.    }   \\

 \hlinewd{1pt}
 \end{tabular}
 \caption{{\sffamily\small Multimodal interactions in the streetscape.}}
 \label{multimodal_interactions}
\end{center}
\end{table}
 }

 \textbf{\sffamily Visuospatial Complexity} \quad    Visual complexity and its effect on visual attention and cognition has been investigated by different disciplines - including cognitive science, psychology, computer science, marketing \cite{Cavalcante2014}. Several definitions have been proposed as well as several methodologies to measure it or measure specific attributes of it (e.g. visual saliency, spatial frequency, subband entropy) \cite{Henderson2009, Madan2018, Rosenholtz2007}. Visual complexity has been broadly defined as the level of detail and intricacy contained within an image or a scene \cite{Snodgrass1980}. With the term visuospatial complexity we consider the combination of visual and spatial characteristics that both coexist in dynamic naturalistic scenes where a person acts. In addition to the visual related features such as colour, contrast, number of objects etc., we also study the size and the structure of the space, the connection of its parts, etc.  Visual attention is being controlled involuntary by external stimulus features (exogenous) such as luminance, and colour; as well as by voluntary internal cognitively relevant features of the world (endogenous) such as people, objects \cite{Carrasco2011}.  Visual search refers to an intentional look in the scene and the visuospatial complexity of the scene is highly correlated to visual search performance and the visual attention patterns as they evolve over time \cite{Eckstein2011}. However,  it is challenging to separate the bottom-up from the top-down cognitive processes (such as visual search, attentional priming, foraging) throughout a naturalistic visual attention task in driving, and therefore to establish which dimension of complexity (low-level scene properties, or high-level semantic properties) affect the visual attention patterns \cite{Schutt2019, Awh2012, Williams2019, Kristjansson2020, Kristjansson2019}. For this reason it is necessary to establish a taxonomy of attributes and parameters (pertaining to visuospatial complexity) quantifying the levels of complexity reliably based on their effects on people's performance and visual attention patterns.

\section{\sffamily Visuospatial Complexity in Everyday Driving: A Human-Centred Model} 

Developing a taxonomy and model of visuospatial complexity necessitates the identification of objective physical attributes that affect visuospatial perception and cognitive functions, such as \emph{visual search} in everyday activities (e.g., driving, walking, cycling). However, the majority of existing empirical evidence is based on studies focusing on static real-world scenes, or abstract shapes and symbols \cite{Forsythe2009, Harper2009, Gartus2017}. For our taxonomy of attributes, we focus on naturalistic dynamic scenes, and we take into consideration recent work about attentional synchrony on dynamic real-world scenes demonstrating the effect of the dynamic attributes on visual attention \cite{Smith2013, Mital2011}. We categories the attributes into {\small\sffamily\textbf{(A1--A3)}}:

{
\begin{table}[t]
\renewcommand{\arraystretch}{1.1}
\begin{center}
 \scriptsize\sffamily
\begin{tabular}{>{\columncolor[gray]{0.92}}l p{12.5cm}}
\hlinewd{1pt}
\rowcolor[gray]{.92}\textbf{VISUOSPATIAL COMPLEXITY} & \textbf{Description} \\\hline\hline

\textbf{Quantitative attributes} &  \\
\hline

 SIZE & [Width / Depth / Height]  The dimensions of the physical space, the area coved by the visual stimulus.  \\[6pt]
 
  CLUTTER   &  \\
 Quantity   &  No. components  (objects, people, shapes etc.) \\
 Variety of Colours & No. colours \\
 Variety of Shapes/Objects & No.  shapes / objects   \\ 
 Objects Density &  No. objects in a defined area \\
 Edges Density &  No. edges of objects in a scene / visual area  \\
 Luminance &  Amount of light emitted / reflected from the scene \\
Saliency &  Particularly prominent objects based on characteristics of colour, luminance and contrast   \\ 
Target-background similarity & Compare similarity in luminance, contrast, structure, or spatial and orientation information \\

\hline
\textbf{Structural  attributes}     &  \\
\hline

Repetition & Recurrence of the same element of group of elements or characteristics on a line, a grid  or a patterns in space \\
Symmetry & Resilience to transformation and movement. Types: reflectional, rotational, translational, helical, fractal   \\ 
Order & Organised elements based on a recognised structure, Varies from poorly organised to highly organised \\
Homogeneity/Heterogeneity  & The state of being all the same kind/ diverse. Varies from single shape repeated to multiple distinct shapes  \\  
Regularity  & Variations in a placement rule across a surface or line; Varies from simple polygons to abstract shapes  \\
Openness & The ratio between empty and full space  \\ 
Grouping & No. elements that are part of a group  \\

\hline
\textbf{Dynamic  attributes}  &   \\
\hline

Motion & No. people or objects moving in the scene \\ 
Flicker   &  Abrupt changes over-time (in luminance, colours, etc.)  \\
Speed | Direction & The rate of change of position with respect to time | Move or facing towards \\
 
\hlinewd{1pt}
\end{tabular}
\caption{{\sffamily\footnotesize Taxonomy of attributes for Visuospatial Complexity}}
\label{tbl:visual complexity model}
\end{center}
\end{table}%
}

\smallskip

\textbf{\small\sffamily {\color{blue!80!black}A1}.\quad Quantitative Attributes.} Objective environmental factors -- referring to low-level  (\emph{lines, edge, contour}) and middle-level (\emph{corners, orientation}) features of the scene -- and their relation to visual perception and cognition, have been a topic of research on computational image processing \cite{Bravo2008, Semizer2019}. Psychophysical and neurophysiological research has shown that these attributes are part of early human visual processing \cite{Smith2013}. Moreover, the physical space and its functional \emph{clutter} are general properties of all scenes that are immediately accessible to humans \cite{Park2014}. \emph{Clutter} refers to overabundance of information, including the \emph{number} of components of the scene, their \emph{variety}, as well as the \emph{density} of the viewed entities from the current perspective. Additionally, \emph{luminance} and \emph{colour}, and their respective retinotopic gradients (contrast), are typically employed in computational models of \emph{saliency}. Similarities between a target with the background in various low-level features or in the spatial arrangement also correlates with visuospatial complexity \cite{Rosenholtz2007}.

\smallskip

\textbf{\small\sffamily {\color{blue!80!black}A2}.\quad Structural Attributes.}\quad   Structural attributes refer to the relations that the elements form due to positioning in space, or the overall distribution in the viewing scene. Many times structural attributes of the scene can mitigate the effects of visuospatial complexity increasing due to low-level quantitative attributes. High regular arrangements of elements in space or in the scene are related to low visuospatial complexity, and more randomised arrangements contribute to higher complexity levels \cite{Doyon-Poulin2012}. \emph{Repetition, symmetry, order}, and \emph{grouping} are related to a decrease in spatial complexity \cite{vanderHelm2000,Salingaros2014}. Likewise, \emph{heterogeneity} (e.g., a single shape repeated vs. multiple distinct shapes), \emph{regularity} of shapes and objects (e.g., simple polygons vs. more abstract shapes), \emph{openness} (i.e., relation between full and empty space) and their relationship to complexity have been investigated in the fields of urban design and architecture \cite{Boeing2018,Salingaros2014}.
 \smallskip

\textbf{\small\sffamily {\color{blue!80!black}A3}.\quad Dynamic Attributes.} \quad Studies of visual attention on videos reveal a significant impact of dynamic features of the scene such as \emph{motion} and \emph{flicker} on top-down as well as bottom-up cognitive processes. Even the same environment can be perceived as more complex due to an increase in velocity of the observer. Taking into consideration dynamic aspects in scene analysis promotes gaze allocation prediction, as previous studies suggest that dynamic cues are predominant, and reliable predictors for greater number of people \cite{Mital2011}. Concerning top-down processes, clusters of attention often coincide with semantically rich objects such as eyes, hands, vehicles, etc. On the other hand, cortical analysis shows a selective response to moving elements on the scene, meaning that we are able to notice moving objects even if we are not looking for them \cite{Rosenholtz1999}. \emph{Flicker}, i.e.,  abrupt changes in luminance over time, has been shown to pop-out independently of observers attention. Moreover, \emph{speed} and \emph{direction} are related to motion but they have different dependences on the way they are encoded during a cognitive task  \cite{Carrasco2011,Mital2011}. 

\smallskip

The role of visuospatial complexity on visual attention depends on the kind of stimuli employed and on the way in which visuospatial complexity is defined, manipulated, and measured. Consequently, there are more significant factors influencing the role of visuospatial complexity on human behaviour that we do not consider in this paper but we cannot ignore. The nature of the \emph{task} or aspects of \emph{familiarity, working memory}, as well as \emph{previous knowledge} of the context, are some them. For instance, semantic grouping of the scene based on previous knowledge during driving indicating where targets expected to appear.  Additionally, studying dynamic scenes reveals the significance of \emph{time} on visual attention, and a continuous interplay between low-level and high-level cognitive functions that affect the visual search performance. Analysing visual attention over time shows that first fixations are more influenced by environmental properties, but after the first $200$ms fixation locations are determined predominantly by top-down processes  \cite{Schutt2019}.

\begin{figure}[t]
\includegraphics[width=0.98\textwidth]{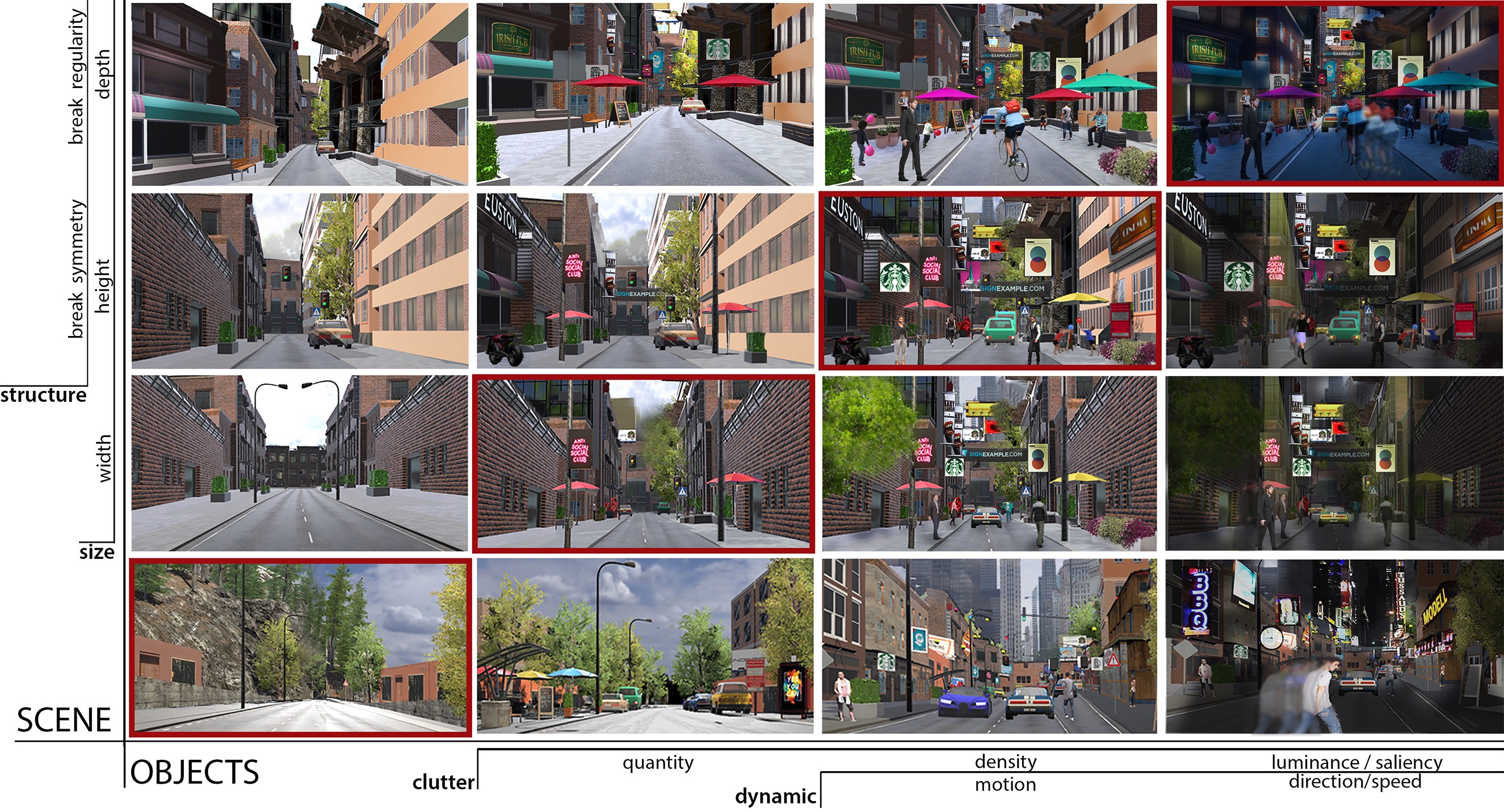} 
\caption{\footnotesize\sffamily Sample of  VR scenes that varies on the level of Visuospatial Complexity. The scenes are developed based on one urban environment with the combining of a number of attributes provided in Table \ref{tbl:visual complexity model}, and it is not an exhaustive representation of the possible outcomes. The scenes are organised in a matrix as follows: the attributes that shape the scene itself are listed on the vertical axis, while the attributes related to objects are introduced on the horizontal axis. In every new line in both direction a new attribute is introduced in addition to the existed ones. The combination provides a new scene. The marked scenes are used for the empirical study of Section \ref{study}.}
\label{matrix}
\end{figure}

\smallskip

\begin{figure}[t]
\begin{center}
\includegraphics[width=0.98\textwidth]{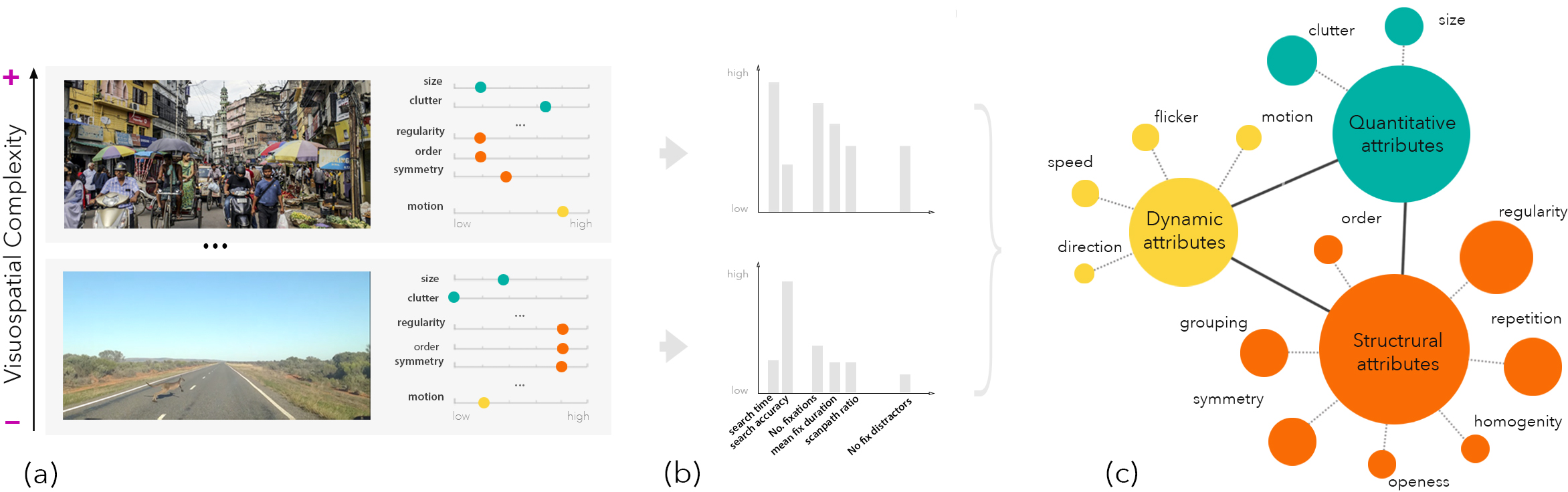}
\caption{\sffamily\footnotesize Visuospatial complexity model: \textbf{(a)} In the top row a high complex scene, and in the bottom row a low complex one, accompanied by the degree of the involved attributes; \textbf{(b)} behavioural metrics from the empirical study; \textbf{(c)} the bubble diagram representing the dynamic relations between the attributes of visuospatial complexity based on the effect they have on human behaviour.}
\label{modelling_process}
\end{center}
\end{figure}

\textbf{\sffamily Towards a model of visuospatial complexity.}\quad   Visuospatial complexity is a function of not only individual factors but also interactions between them. For instance, clutter can increase complexity, however high complexity may also be achieved with a combination of low clutter and low contribution of structural attributes (e.g. order, heterogeneity). A systematic analysis of different combinations of attributes can provide a better understanding of the aggravation or counterbalance dynamics between the attributes and their effect on human behaviour. To empirically define a model of visuospatial complexity for dynamic naturalistic scenes, we use the taxonomy introduced in Table \ref{tbl:visual complexity model} to develop a number of scenes in virtual reality (VR) that differ on the combination of attributes involved, as well as on the degree of each attribute the scene contains (Fig. \ref{modelling_process}a).\footnote{An important aspect not considered in this paper is high-level \emph{event perception} \cite{Tversky2013}; presently, work is also in progress to include high-level event segmentation primitives complementing the developed visuospatial complexity model.}
 This way we create a matrix of possible scenes (Fig. \ref{matrix}), and use a number of them as the dynamic stimulus for our empirical study. The matrix provides an indication for a scale of visuospatial complexity based on the current knowledge from the literature on the effect of individual attributes on human behaviour. However, the empirical result of our study will be used to confirm or decline this indication, and further investigate the \emph{weight} of each attribute to the overall effect on human behaviour (Fig. \ref{modelling_process}b). To represent these interactions between the attributes and their contribution to the overall complexity level we are developing a dynamic bubble diagram (Fig. \ref{modelling_process}c) that demonstrates the correlation coefficients of visuospatial complexity attributes and the behavioural metrics. We expect a stronger correlation between the visual search inefficiency and factors of clutter than size or structural attributes. Additionally, we expect quantitative attributes to have a negative correlation with visual search performance, while the structural ones to have a positive correlation. Although this effect can be counterbalanced.

\section{\sffamily  Visual Search in the Streetscape: An Ongoing Empirical Study} \label{study}

 \textbf{\sffamily  A visual search task in driving.} \quad Driving performance has been shown to depend on visuospatial complexity in this case the streetscape, but also on the complexity of the driving task, and other factors that affect cognitive resources, such as individual differences, fatigue, age, dual task requests, etc \cite{Edquist2009, Wierda1996}. We are primarily concerned about the extent of which environmental factors of the dynamic scene affect the driving performance, such as locating and responding to traffic, directions or warning signs, hazard detection, and vehicle control etc. Theses actions are closely connected to visual perception and visual search processes. Visual search in driving primarily refers to detection of signs, obstacles, or road users, as well as situation awareness, including  perception of hazard.

\smallskip 

\begin{figure}[t]
 \begin{center}
\includegraphics[width=0.94\textwidth]{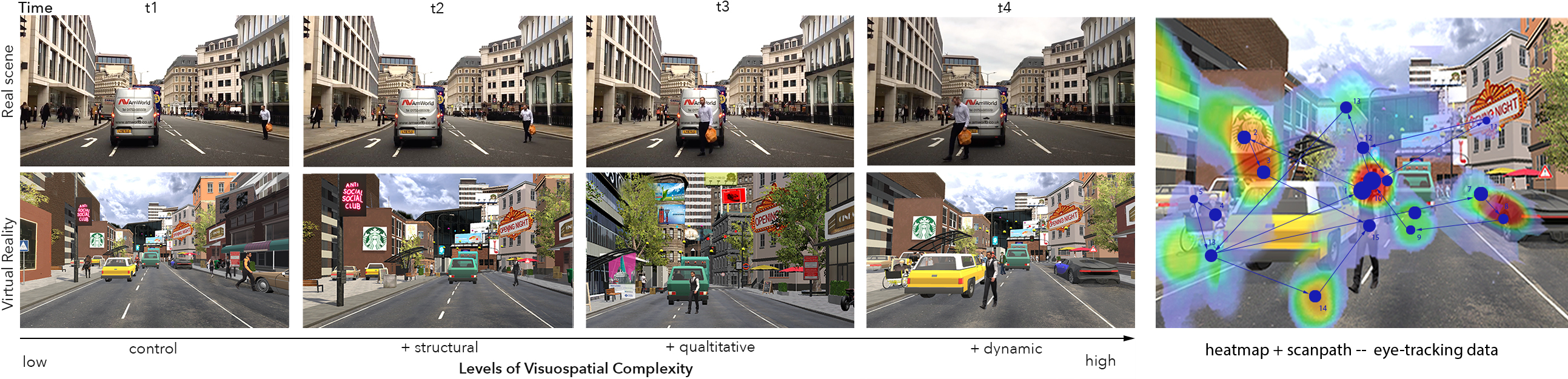}
 \caption{\sffamily\footnotesize Example scenarios as it evolves over time (t1-t4), in the real scene (top) and in a virtual scene (bottom). The development of the VR scenes in four levels of complexity are demonstrated on the horizontal axis (based on Figure \ref{matrix}). Starting from a control scene, in every new scene towards right, new attributes from a new category (\ref{tbl:visual complexity model}) are introduced in addition to the existed ones. On the right, a sample heatmap of eye-tracking data for one VR scene.}
 \label{scenario1}
 \end{center}
 \end{figure}

 \textbf{\sffamily  Four scenarios in four levels of visuospatial complexity.} \quad We collect and analyse a list of scenarios with dynamic scenes in streetscape that differ on the events and the interactions between the street users involved. The scenarios are collected from online video footage, following the indications about hazard situations published on Accidence Research report by the German Insurance Association \cite{GDV2017}, and categorised based on the nature of the incident (Table: \ref{multimodal_interactions}).  For the behavioural study in progress we choose four scenarios that differ on the multimodal interaction event and the practical actions used (Table: \ref{multimodal_interactions}), and we replicate them in VR environment, in four different levels of visuospatial complexity. One example scenario on \emph{inattentive crossing by pedestrian} is presented in Fig. \ref{scenario1}.  It is a typical scenario of pedestrian on a low traffic street with no official indication for crossing such as zebra crossing or traffic light. The pedestrian who is crossing more than one lane, firstly deviates from his path on the sidewalk and then he performs a slow speed crossing, while he is altering his visual attention states multiple times, between monitoring traffic \timePoint{1}, crossing the street being inattentive \timePoint{2}, sharing attention with a driver \timePoint{3}, and continue to cross the next lane being visually inattentive \timePoint{4}. The visual search task for the driver is to detect the pedestrian and elaborate on his behaviour. The pedestrian - considered from the driver's point of view as the dynamic visual target - passes gradually from the peripheral view of the driver to the fovea area, while he is changes orientation and speed.  
\smallskip

As the visual search process can be divided in different stages (early phase of search guidance, verification component), no single metric can describe the effect of visuospatial complexity on a search process entirely; rather it a combination of several metrics for performance and physiological measurements such as eye-tracking which can provide a more precise assessment of the effect of visuospatial complexity on human behaviour.  We collect eye-tracking data from the participants, as well as reaction time, and accuracy to the target (Table \ref{metrics})\footnote{The study is conducted using HTC Vive Pro Eye headset with embedded eye-tracking device. Unity Game Engine was used for the VR simulation of the scenes. VR is currently firmly established as an experimental tool; it combines a high degree of control with high level of ecological validity, and has shown important benefits for basic psychology and neuroscience research, especially on aspects of visual perception and  spatial cognition \cite{Bohil2011,Olk2018,Scarfe2015,Wilson2015}. }.

 For the analysis we define areas of interest and distractors according to the nature of the scene and the interactions involved. We further combine the results from the physiological and performance measurements and use them to trace the effect of visuospatial complexity levels. We expect the combination of measurements to provide  insights about the attentional cost and the underline cognitive processes affected by visuospatial complexity. 
\smallskip

 \smallskip

\section{\sffamily Outlook} \label{outlook}
The road ahead (in autonomous driving) presents several opportunities:  next-generation  developments to achieve (legal) deployment and user / ethical acceptability require attention to a wider spectrum of autonomous driving challenges from the viewpoint of human-centred technology design encompassing natural human-machine interaction, industry-wide benchmarking, standardisation and statutory validation of technology components.  The work reported in this paper is being conducted in synergy with research in computational \emph{cognitive vision} \cite{BhattECAI20}, particularly  in relation with the development of integrated vision and semantics solutions for active, explainable visual sensemaking for autonomous vehicles \cite{SuchanECAI20,SuchanIJCAI19}. The empirically founded model developed in this paper also constitutes the basis for computational assessment and analysis of visuospatial complexity. Towards this, we use the empirical results to develop a declaratively grounded computational model integrating KR and vision (\cite{CogSys-Symmetry-2018,DBLP:conf/ilp/SuchanBS16}) towards: \textbf{a)} learning behavioural models from empirical data, i.e., using (statistical) relational learning to induce weighted dependencies and constraints within the model, and \textbf{b)} utilising the computational model to examine and construct benchmarking datasets ensuring variety in driving conditions, naturalistic multimodal human interaction events, challenging and diverse environments, etc. Out goal is that a computational model of visuospatial complexity may provide a measure for the complexity of benchmarking datasets/scenarios and thus serve as a guideline to develop realistic human-centred evaluation criteria. We posit that such interdisciplinary studies are needed to better appreciate the complexity and spectrum of varied human-centred challenges in autonomous driving.

 {
 \begin{table}[t]
 \renewcommand{\arraystretch}{1.1}
 \begin{center}
 \scriptsize\sffamily
 \begin{tabular}{>{\columncolor[gray]{0.92}}l l}
 \hlinewd{1pt}
\rowcolor[gray]{.92} \textbf{METRICS} & \textbf{Description -- Correlation with visuospatial complexity} \\\hline\hline
 
 \textbf{Performance Evaluation}     &  \\[0.2 pt] 
 \hline
 
 Search time  & domain specific measurement related to the task  (e.g. detection of   \\
 Accuracy  & pedestrians, detection of objects) \\
 
 \hline
 \textbf{Physiological measurements -- Eye-tracking}     &  \\
 \hline
 
 Latency of first saccade & indicates confusion and uncertainty\\
 
 Total number of fixations &  large number indicates decreased search efficiency \\
 
 Number of fixations on distractors in AOIs & larger when the target's location is in high complex area  \\
 
 Number of fixations excluding those on distractors & indicate that other aspects like high variation contribute to complexity \\
 
  Scanpath ration & big length of scanpath indicates poor search efficiency   \\
 
 Mean fixation duration &  longer fixation duration indicates difficulty in processing information \\
 
 Final saccade length & indicates how easily noticeable the target is in the peripheral vision \\
 
 \hlinewd{1pt}
 \end{tabular}
 \caption{{\sffamily\footnotesize Summary of metrics used for visual search evaluation and assessment of visuospatial complexity effects}}
 \label{metrics}
 \end{center}
 \end{table}%
 }

\footnotesize
\linespread{0.9}
\bibliographystyle{ecai}

\end{document}